\documentclass[10pt,twocolumn,letterpaper]{article}

\usepackage[pagenumbers]{cvpr} % To force page numbers, e.g. for an arXiv version

\usepackage[ruled,vlined]{algorithm2e}
\usepackage{multirow}
\usepackage{indentfirst}
\definecolor{cvprblue}{rgb}{0.21,0.49,0.74}
\definecolor{mydarkgreen}{RGB}{0, 128, 0}
\usepackage[pagebackref,breaklinks,colorlinks,citecolor=cvprblue]{hyperref}

\def\paperID{4130} % *** Enter the Paper ID here
\def\confName{CVPR}
\def\confYear{2024}

\title{Dynamic Graph Representation with Knowledge-aware Attention for Histopathology Whole Slide Image Analysis}

\author{Jiawen Li\textsuperscript{1*}, Yuxuan Chen\textsuperscript{1*}, Hongbo Chu\textsuperscript{1*}, Qiehe Sun\textsuperscript{1}, Tian Guan\textsuperscript{1}, Anjia Han\textsuperscript{2\dag}, Yonghong He\textsuperscript{1\dag}\\
\textsuperscript{1}Shenzhen International Graduate School, Tsinghua University\\
\textsuperscript{2}Department of Pathology, The First Affiliated Hospital of Sun Yat-sen University\\
{\tt\small \{lijiawen21, chenyx23, zhu-hb23\}@mails.tsinghua.edu.cn}\\{\tt\small hananjia@mail.sysu.edu.cn}, {\tt\small heyh@sz.tsinghua.edu.cn} \\
}

\begin{document}
\maketitle
\begin{abstract}

Histopathological whole slide images (WSIs) classification has become a foundation task in medical microscopic imaging processing. Prevailing approaches involve learning WSIs as instance-bag representations, emphasizing significant instances but struggling to capture the interactions between instances. Additionally, conventional graph representation methods utilize explicit spatial positions to construct topological structures but restrict the flexible interaction capabilities between instances at arbitrary locations, particularly when spatially distant. In response, we propose a novel dynamic graph representation algorithm that conceptualizes WSIs as a form of the knowledge graph structure. Specifically, we dynamically construct neighbors and directed edge embeddings based on the head and tail relationships between instances. Then, we devise a knowledge-aware attention mechanism that can update the head node features by learning the joint attention score of each neighbor and edge. Finally, we obtain a graph-level embedding through the global pooling process of the updated head, serving as an implicit representation for the WSI classification. Our end-to-end graph representation learning approach has outperformed the state-of-the-art WSI analysis methods on three TCGA benchmark datasets and in-house test sets. Our code is available at \href{https://github.com/WonderLandxD/WiKG}{https://github.com/WonderLandxD/WiKG}.
\end{abstract}
{\let\thefootnote\relax\footnote{* Contributed equally. \dag Corresponding authors.}}
\section{Introduction}
\label{sec:intro}

Pathology is the gold standard for disease diagnosis. Traditional histopathological diagnostics are performed manually by examining pathological slides using an optical microscope. However, due to social concerns such as environmental changes and an increasing aging population, those diagnostic approaches are facing significant challenges  \cite{ellison2018impact}. With the advancement of digital scanning technologies, physical slides can now be converted into high-resolution WSIs that preserve the complete pathological tissues. This advancement has markedly improved the work efficiency of pathologists \cite{pell2019use} and stimulated demand for intelligent WSI diagnostic tools \cite{madabhushi2009digital,madabhushi2016image,pallua2020future}.
\begin{figure}[tbp]
\centerline{\includegraphics[width=0.95\columnwidth]{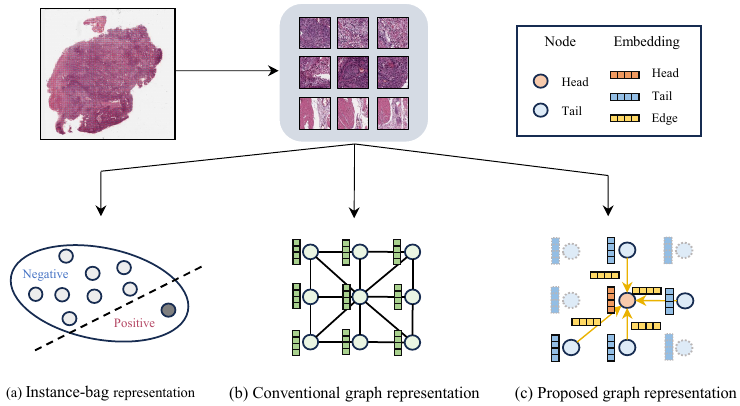}}
\caption{Illustration of three different frameworks of instance-bag representation, conventional undirected graph representation, and the proposed dynamic graph representation.}
\label{intro}
\end{figure}

In recent years, many studies have shown promising results by continuously exploring deep learning in WSIs. However, the professional interpretation of WSIs and their large-scale data type make obtaining manual annotation at the pixel level difficult. To alleviate this issue, researchers have developed weakly supervised algorithms that can be trained using only slide-level labels \cite{sharma2021cluster,wang2019weakly}. Many existing weakly supervised studies on WSI analysis are based on embedding-based multiple instance learning (MIL) \cite{maron1997framework}, which partitions WSIs into patches (instances) and aggregates their embedding information. However, MIL overly emphasizes the impact of patches on the global representation but ignores the interconnections. Such interconnections often indicate the potential development trends of tumors and provide pathologists with deeper relational insights, contributing to the construction of tumor microenvironments. For instance, the interactions between tumor cells and inflammatory cells assist in analyzing tumor staging \cite{chan2023histopathology} and clinical prognosis \cite{chen2021whole, lee2022derivation}. Graph Neural Networks (GNNs) are emerging as promising tools for WSI analysis, primarily because they focus more on local similarities within the entity topology \cite{ahmedt2022survey}. GNNs represent WSIs as graphs, treat patches as nodes, and transfer and aggregate feature information of patches through the interaction between nodes to obtain graph-level representations.
%GNNs represent WSIs as graphs, with patches as nodes, and update the features of patches through node interactions to generate graph-level representations. 

Graph-based WSI analysis methods are currently devoted to modeling through the spatial positional relationships of patches. While these approaches have achieved some success, they exhibit some deficiencies. (1) Explicit spatial topology restricts the ability of distant entities to explore information with one another. (2) GNNs for WSIs are typically modeled as undirected graphs, which overlooks the directed contributions of adjacent entities. (3) The over-parameterization of GNNs in WSI analysis may lead to over-fitting, which reduces the generalization ability.

For this purpose, we introduce a dynamic graph representation based on knowledge-aware attention called WiKG, which fully leverages the interactions among entities to analyze WSIs. Specifically, we parameterize the head and tail embeddings for each patch, which are instrumental in modeling the directed graphs of WSIs (Figure \ref{intro} shows the formal differences between proposed directed graph representation and other methods). Following this, we design edge embeddings based on the interactions between the heads and tails, and construct a knowledge-aware attention mechanism that aggregates neighbor information. This mechanism injects the embeddings of head, tail, and edge triplets into neighbor knowledge attributes, allowing more valuable entities to propagate more practical information. To demonstrate the effectiveness of our method, we conduct extensive evaluations on three public benchmark TCGA datasets, including typing and staging for esophageal carcinoma, kidney cancer, and lung cancer. In addition, we collect in-house frozen section lung cancer WSIs for external testing to showcase the generalization ability of our method. Through comparisons with various state-of-the-art WSI analysis methods and ablation studies, we demonstrate the superior performance of WiKG in histopathological WSI analysis tasks.
\begin{figure*}[ht]
\centerline{\includegraphics[width=2\columnwidth]{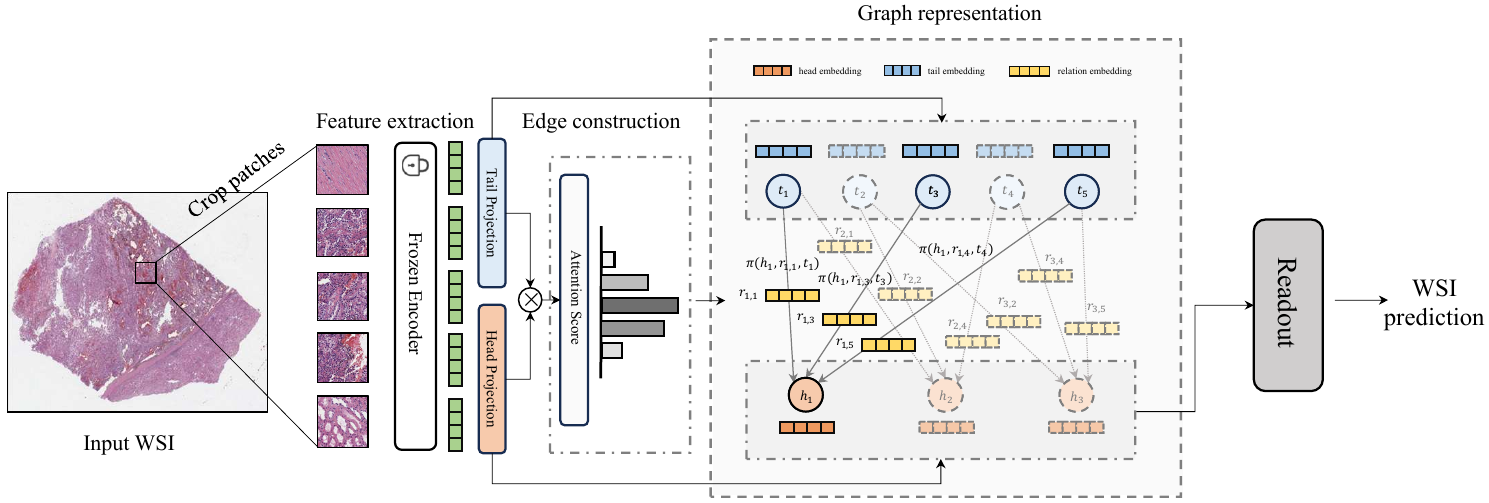}}
\caption{The framework of our proposed method for WSI analysis, including patch feature extraction, dynamic edge construction based on head and tail embeddings, graph representation learning, and the prediction of WSIs.}
\label{main}
\end{figure*}

\section{Related Works}
\subsection{Application of MIL in Histopathology WSIs}
Given the consistency between MIL theory and the diagnosis process of pathologists, the development of MIL algorithms for WSI analysis has become a principal research trend \cite{gadermayr2022multiple}, where models are constructed in an instance-bag format. First, researchers attempt to generate WSI predictions based on instance scores \cite{campanella2019clinical,hou2016patch,lerousseau2020weakly,xu2019camel}. As the embedding-based MIL method has developed, the research trend gradually turned to it because of its more stable convergence and better performance. Conventional embedding-based MIL approaches involve compressing patches into embeddings and aggregating these to learn WSI-level features for final predictions \cite{ilse2018attention,lu2021data,li2021dual,chen2023dmil}. For example, ABMIL \cite{ilse2018attention} learns an attention score for each instance and performs a weighted aggregation to obtain a bag representation at the WSI level. Lu \textit{et al.} \cite{lu2021data} propose an improved ABMIL by introducing clustering-constraint loss and single/multiple-branch attention to enhance WSI classification capabilities. Li \textit{et al.} \cite{li2021dual} propose a dual-stream network to capture the latent relationships between the most suspicious positive instances and others. Shao \textit{et al.} \cite{shao2021transmil} integrate transformer networks into the MIL paradigm to update inter-instance embeddings efficiently. Zhang \textit{et al.} \cite{zhang2022dtfd} build upon ABMIL by introducing the concept of pseudo-bags to enhance the attention information and proposed a feature distillation strategy to fuse representations at the pseudo-bag level. Though these methods have endeavored to improve the analytical performance of WSIs, they still lack representation of the informative interactions to inter-patch interactions.

\subsection{Graph Representation in Digital Pathology}
The tumor microenvironment, as delineated by histological phenotypes and topological distribution, plays a pivotal role in pathological diagnosis. Graph-based deep learning approaches, outstanding in encoding representations and capturing intra- and inter-nodal interactions, have attracted widespread attention in digital pathology \cite{ahmedt2022survey}. Specifically, Zhou \textit{et al.} \cite{zhou2019cgc} and Chen \textit{et al.} \cite{chen2020pathomic} construct edges through the spatial position of cells, coupled with hand-crafted features to form the cell graph representations in tissue. As the paradigm of efficient feature extraction through deep learning, training neural networks at the cellular level can enhance the performance of the cell graph \cite{pati2022hierarchical}. Some studies have also designed topological structures by integrating representations of tissue regions and cells. For example, HACT-Net \cite{pati2020hact} combines tissue and cell graph representations to classify histopathological regions of interest in the breast subtype. SHGNN \cite{hou2022spatial} constructs a spatially hierarchical GNN framework for joint representation learning on tissue and cell structures.

With the necessity to analyze whole slide images (WSIs) in tissue pathology, graph representations that use patches as entities have also been widely discussed. For instance, Chen \textit{et al.} \cite{chen2021whole} model the local and global topological structures within the tumor microenvironment by hierarchically aggregating patch-level histological features. Zheng \textit{et al.} \cite{zheng2022graph} combine the spatial representation of WSIs with Vision Transformers to predict disease grades. Hou \textit{et al.} \cite{hou2022h} develop a multiscale hierarchical representation using heterogeneous graphs with resolution attributes to explicitly model the spatial scale relationships of patches. Zhao \textit{et al.} \cite{zhao2023mulgt} establish a multi-task learning framework for WSI analysis using graph neural networks and Transformers as the building base. Chan \textit{et al.} \cite{chan2023histopathology} analyze WSIs by representing them as heterogeneous graphs, utilizing the relationships between different types of nuclei. While these methods have demonstrated impressive results, most rely on fixed spatial constructs for edge creation, limiting the ability to explore mutual information freely. Simultaneously, they fail to fully account for the directional complexity of contextual interactions between patches.

\section{Methodology}
In this section, we introduce our proposed WiKG, including how dynamic graph construction works and the knowledge-aware attention mechanism for updating node features. Figure \ref{main} presents our proposed framework.
\subsection{Dynamic Graph Construction}
Unlike conventional methods that construct graphs using spatial relationships \cite{chen2021whole,zheng2022graph,hou2022h}, our approach is dedicated to quantifying positional relationships between patches through learnable implicit features. For a given WSI, we first use the Otsu threshold algorithm \cite{otsu1979threshold} to distinguish the foreground tissue regions and apply the sliding window operation to divide them into non-overlapping patches $X = \{ x_1,x_2,\dots,x_n\}$, which are denoted as graph nodes. Subsequently, we utilize a feature encoder $f$ (e.g., Vision Transformer \cite{dosovitskiy2020image} pretrained on ImageNet \cite{deng2009imagenet}) to obtain embeddings for each patch. Following this, we project these into head and tail embeddings using two separate linear projection layers, where the head aims to explore the correlation between other patches and itself, and the tail aims to explore its contribution to other patches.%, which is represented as: %For a given WSI, we begin by distinguishing the foreground tissue regions using the Otsu thresholding algorithm\cite{otsu1979threshold} within a sliding window pattern, partitioning them into non-overlapping patches $X = \{ x_1,x_2,\dots,x_n\}$, which are denoted as graph nodes. 
\begin{equation}
    h_i = W_{h} \space f(X)\text{, \space\space}  t_i = W_{t} \space f(X),
\end{equation}
where $h_i$ and $t_i$ respectively denote the head and tail embeddings of patch $i$. We then compute the dot product of these and employ a softmax function to quantify the similarity between heads and tails, which is expressed as:

\begin{equation}
    \omega_{i,j} = \frac{h_i^\mathrm{T}t_j} {{\textstyle \sum_{j=1}^{N}(h_i^\mathrm{T} t_j)}}, 
\end{equation}
where $\omega_{i,j}$ represents the similarity score between the head of patch $i$ and the tail of patch $j$. For each patch $i$, the top $k$ patches with the highest similarity score are selected as the neighbors of patch $i$, which is described as:
%Next, we record the top-k connections as neighbors of patch $i$, which is described as: 
\begin{equation}
    \mathcal{N} (i) = \{ j \in V : \omega_{i,j} \in \mathrm{Topk}\{\omega_{i,j}\}_{j=1}^N \},
\end{equation}
where $V$ represents the patch set, and $|V| = N$, $|\mathcal{N} (i)| = k$. Our topological structure also assigns embeddings for directed edges, obtained using head and tail embeddings:
\begin{equation}
    r_{i,j} =  \omega_{i,j} t_j + (1 - \omega _{i,j}) h_i \text{, \space\space for every $j \in \mathcal{N}(i)$}, 
\end{equation}
where $r_{i,j}$ represents the edge embedding from patch $j$ to patch $i$. Through the above operations, we delineate a WSI as a dynamic graph representation $G = (V, \mathcal{E}, \mathcal{F}, \mathcal{R})$, where $V$ denotes the set of nodes, $\mathcal{E}$ denotes the set of edges, $\mathcal{F}$ denotes the set of head and tail embeddings and $\mathcal{R}$ denotes the set of directed edge embeddings. $\text{{\Large$\varepsilon$}} = \{ (h,r,t): (h,t) \in \mathcal{F}, r \in \mathcal{R} \}$ introduces the triplet of head, tail, and their high dimensional relation on each directed edge. Algorithm \ref{algo:construction} provides the pseudo-code for our proposed dynamic graph construction in a Pytorch-like style.

\subsection{Knowledge-aware Attention Mechanism}
\definecolor{commentcolor}{RGB}{110,154,155}   % define comment color
\newcommand{\PyComment}[1]{\ttfamily\textcolor{commentcolor}{\# #1}}  % add a "#" before the input text "#1"
\newcommand{\PyCode}[1]{\ttfamily\textcolor{black}{#1}} % \ttfamily is the code font
\begin{algorithm}[t] \scriptsize
\SetAlgoLined
    \PyComment{input: patch embedding (B, N, C)} \\
    \PyComment{output: relation embedding (B, N, k, C)} \\
    \PyComment{B: batch size} \\
    \PyComment{N: patch number} \\
    \PyComment{C: feature dimension} \\
    \PyComment{K: neighbor nodes number} \\
    \hspace*{\fill} \\
    \PyComment{linear projection of head, tail} \\
    \PyCode{e\_h, e\_t = linear\_ht(input).chunk(2, dim=-1)} \\
    \hspace*{\fill} \\
    \PyComment{similarity of head and tail} \\
    \PyCode{attn\_logit = (e\_h * scale) @ e\_t.transpose(-2, -1)} \\
    \hspace*{\fill} \\
    \PyComment{chooses top k weight and index} \\
    \PyCode{topk\_weight, topk\_index = torch.topk(attn\_logit, k=k, dim=-1)} \\
    \hspace*{\fill} \\ 
    \PyComment{expand topk\_index to match dimensions} \\
    \PyCode{topk\_idx\_expand = topk\_index.expand(B, -1, -1)} \\
    \hspace*{\fill} \\ 
    \PyComment{create a range tensor for secondary indexing} \\
    \PyCode{batch\_indices = torch.arange(B).view(-1, 1, 1)} \\
    \hspace*{\fill} \\ 
    \PyComment{generate neighbor features (B, N, k, C)} \\
    \PyCode{Nb\_h = e\_t[batch\_indices, topk\_idx\_expand, :]} \\
    \hspace*{\fill} \\ 
    \PyComment{weight probability} \\
    \PyCode{topk\_prob = F.softmax(topk\_weight, dim=2)} \\
    \hspace*{\fill} \\ 
    \PyComment{generate edge embeddings (B, N, k, C)} \\
    \PyCode{eh\_r = topk\_prob.unsqueeze(-1) * Nb\_h + (1 - topk\_prob).unsqueeze(-1) @ e\_h.unsqueeze(2)} \\
\caption{Directional Edge Construction}
\label{algo:construction}
\end{algorithm}

\begin{figure}[t]
\centerline{\includegraphics[width=0.6\columnwidth]{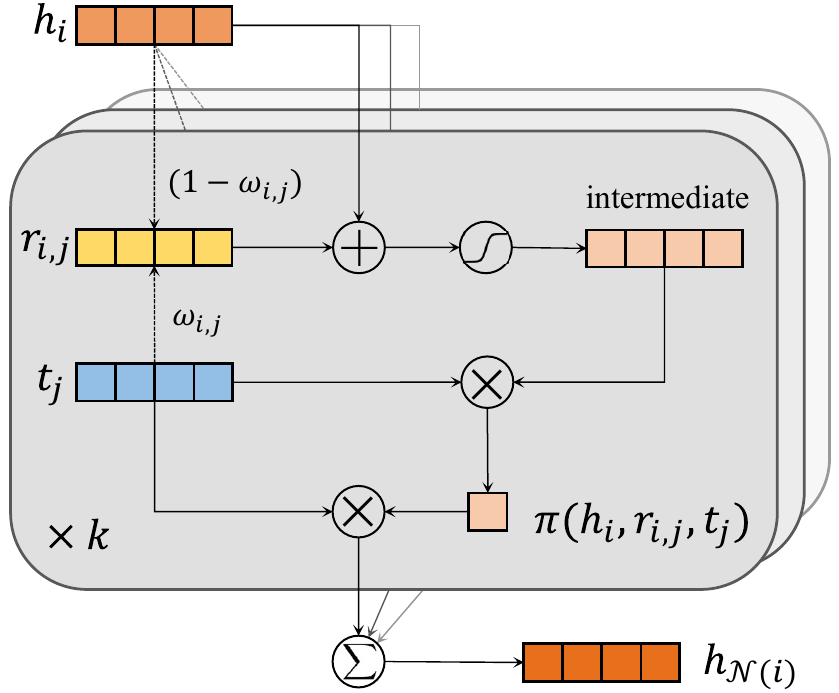}}
\caption{Illustration of our proposed knowledge-aware attention mechanism, including aggregation between head, tail, and edge embeddings.}
\label{edge}
\end{figure}

To take full advantage of the node relationships of the above graph construction, inspired by graph for recommendation \cite{fan2019graph, wang2019kgat}, we propose a knowledge-aware attention mechanism for information propagation and aggregation across nodes. For patch $i$, we compute a linear combination of the tail embeddings of its neighbors $\mathcal{N} (i)$ to characterize its first-order connectivity structure:
\begin{equation}
    h_{\mathcal{N} (i)} = \sum_{j \in \mathcal{N} (i)} \pi(h_i,r_{i,j},t_j)t_j,
\end{equation}
where $\pi(h,r,t)$ is a weighting factor that guides how much information from each tail will be propagated to the head. We use the nonlinear combination of triples to calculate $\pi(h,r,t)$, which is expressed as follows:
\begin{equation}
    u(h_i,r_{i,j},t_j) = t_j^\mathrm{T} \space \mathrm{tanh}(h_i + r_{i,j}),
\end{equation}
where the hyperbolic tangent $\mathrm{tanh}(\cdot )$ serves as a non-linear activation function, facilitating an appropriate gradient flow encompassing both negative and positive values. This combination permits the assessment of the proximity between tails and heads to reveal differences between neighbors. Subsequently, we employ the softmax function to normalize these combinations:
\begin{equation}
    \pi(h_i,r_{i,j},t_j) = \frac{\exp \{u(h_i,r_{i,j},t_j)\}}{ {\textstyle \sum_{j\in \mathcal{N} (i)} \exp \{u(h_i,r_{i,j},t_j)\}}   }. 
\end{equation}
By modeling relationships of triplets and describing them as knowledge information for edges, head nodes are allowed to measure signals from tail nodes and capture them efficiently. Figure \ref{edge} illustrates the implementation of our proposed knowledge-aware attention mechanism. Ultimately, we fuse the aggregated neighbor information with the original head to form a new head representation. We adopt a dual-interaction mechanism to facilitate the exchange of more messages between nodes:
\begin{equation}
    h_i = \sigma_1 (\mathrm{W_1}(h_i + h_{\mathcal{N}(i)}) ) + \sigma_2 (\mathrm{W_2}(h_i \odot h_{\mathcal{N}(i)})),   
\end{equation}
where $\sigma$ is the activation function, such as $\mathrm{LeakyReLU}$, and $W$ denotes a learnable transformation matrix. Finally, we employ a $\mathrm{Readout}$ function to generate graph-level embeddings and a softmax function to obtain the probability score of the WSI:
\begin{equation}
    \hat{Y} = \mathrm{Softmax}(\mathrm{Readout(G)}),
\end{equation}
where $\mathrm{Readout}$ is a global pooling layer, such as mean or max pooling, and $\hat{Y}$ denotes the predicted probabilities. During training, the cross-entropy loss function is utilized as the objective loss for the WSI classification task, which is expressed as:
\begin{equation}
    \mathcal{L}_{ce} = -\frac{1}{M} \sum_{m=1}^{M}\sum_{c=1}^{C} Y_{m,c} \ln(\hat{Y}_{m,c}),  
\end{equation}
where $C$ is the number of categories, $M$ is the number of training samples, and $Y$ is the one-hot label.

\section{Experiments}
\begin{table*}[htbp]
\centering
\resizebox{2\columnwidth}{!}{
\renewcommand{\arraystretch}{1.5}{
\begin{tabular}{lccccccccc}
\hline
\multirow{2}{*}{\textbf{Method}} &
  \multicolumn{3}{c}{\textbf{TCGA-ESCA}} &
  \multicolumn{3}{c}{\textbf{TCGA-KIDNEY}} &
  \multicolumn{3}{c}{\textbf{TCGA-LUNG}} \\ \cline{2-10} 
 &
  \textbf{Accuracy} &
  \textbf{AUC} &
  \textbf{F1-score} &
  \textbf{Accuracy} &
  \textbf{AUC} &
  \textbf{F1-score} &
  \textbf{Accuracy} &
  \textbf{AUC} &
  \textbf{F1-score} \\ \hline
ABMIL \cite{ilse2018attention} &
  $85.61_{2.00}$ &
  $90.74_{0.90}$ &
  \multicolumn{1}{c|}{$85.70_{1.93}$} &
  $94.97_{0.85}$ &
  $98.69_{0.65}$ &
  \multicolumn{1}{c|}{$94.98_{0.88}$} &
  $80.67_{2.37}$ &
  $84.80_{3.77}$ &
  $80.35_{2.76}$ \\
CLAM-SB \cite{lu2021data} &
  $88.27_{2.30}$ &
  $93.56_{1.40}$ &
  \multicolumn{1}{c|}{$88.28_{2.28}$} &
  $96.68_{0.93}$ &
  $99.49_{0.20}$ &
  \multicolumn{1}{c|}{$96.67_{0.94}$} &
  $82.89_{1.54}$ &
  $89.25_{1.87}$ &
  $82.86_{1.52}$ \\
CLAM-MB \cite{lu2021data} &
  $87.21_{4.05}$ &
  $92.83_{1.64}$ &
  \multicolumn{1}{c|}{$87.27_{4.03}$} &
  $96.92_{0.42}$ &
  $99.55_{0.08}$ &
  \multicolumn{1}{c|}{$96.91_{0.42}$} &
  $80.81_{1.08}$ &
  $88.66_{1.67}$ &
  $80.77_{0.98}$ \\
DSMIL \cite{li2021dual} &
  $87.20_{0.82}$ &
  $91.91_{1.34}$ &
  \multicolumn{1}{c|}{$87.25_{0.79}$} &
  $96.19_{1.31}$ &
  $99.28_{0.26}$ &
  \multicolumn{1}{c|}{$96.17_{1.31}$} &
  $81.80_{1.20}$ &
  $88.30_{1.60}$ &
  $81.78_{1.27}$ \\
TransMIL \cite{shao2021transmil} &
  $86.70_{2.04}$ &
  $92.85_{2.06}$ &
  \multicolumn{1}{c|}{$86.81_{1.94}$} &
  $95.38_{1.47}$ &
  $99.22_{0.51}$ &
  \multicolumn{1}{c|}{$95.36_{1.49}$} &
  $78.31_{2.37}$ &
  $85.89_{3.48}$ &
  $78.03_{2.66}$ \\
DTFD-MIL \cite{zhang2022dtfd} &
  $88.81_{2.49}$ &
  $93.79_{2.22}$ &
  \multicolumn{1}{c|}{$88.88_{2.43}$} &
  $96.27_{1.00}$ &
  $99.41_{0.30}$ &
  \multicolumn{1}{c|}{$96.26_{1.01}$} &
  $82.41_{0.82}$ &
  $88.70_{1.45}$ &
  $82.37_{0.88}$ \\
HIPT \cite{chen2022scaling} &
  $89.24_{4.25}$ &
  $93.79_{3.12}$ &
  \multicolumn{1}{c|}{$89.27_{4.27}$} &
  $96.84_{0.85}$ &
  $99.48_{0.14}$ &
  \multicolumn{1}{c|}{$96.83_{0.86}$} &
  $80.39_{1.13}$ & 
  $86.91_{0.58}$ &
  $80.35_{1.14}$ \\
GTP \cite{zheng2022graph} &
  $78.03_{5.80}$ &
  $84.61_{5.33}$ &
  \multicolumn{1}{c|}{$77.76_{6.09}$} &
  $83.54_{4.35}$ &
  $93.71_{3.33}$ &
  \multicolumn{1}{c|}{$82.76_{5.03}$} &
  $78.07_{1.31}$ &
  $84.01_{2.06}$ &
  $77.98_{1.25}$ \\
Patch-GCN \cite{chen2021whole} &
  $88.30_{2.13}$ &
  $93.98_{0.83}$ &
  \multicolumn{1}{c|}{$88.37_{2.06}$} &
  $96.68_{1.11}$ &
  $99.53_{0.22}$ &
  \multicolumn{1}{c|}{$96.67_{1.10}$} &
   $79.72_{3.67}$ &		
   $87.83_{4.30}$ &
   $79.88_{3.59}$ \\ \hline
\textbf{WiKG (ours)} &
  $\mathbf{90.37_{3.14}}$ &
  $\mathbf{95.23_{2.90}}$ &
  \multicolumn{1}{c|}{$\mathbf{90.40_{3.13}}$} &
  $\mathbf{97.08_{0.71}}$ &
  $\mathbf{99.65_{0.12}}$ &
  \multicolumn{1}{c|}{$\mathbf{97.08_{0.71}}$} &
  $\mathbf{84.02_{0.72}}$ &
  $\mathbf{90.78_{1.24}}$ &
  $\mathbf{83.93_{0.64}}$ \\ \hline
\end{tabular}}}
\caption{Cancer typing results [\%] of various methods on TCGA-ESCA, TCGA-KIDNEY, and TCGA-LUNG datasets.}
\label{typing}
\end{table*}

% In this section, we introduce the performance of the proposed WiKG on three public histopathological benchmark datasets and compare it with other state-of-the-art WSI analysis algorithms. Moreover, we also conduct ablation studies on edge construction, know-aware attention, and the number of neighbor nodes. Finally, we perform additional experiments on the in-house test set to show the generalization of the trained models.

In this section, we introduce the performance of the proposed WiKG on three public histopathological benchmark datasets and compare it with other state-of-the-art WSI analysis algorithms. Moreover, we also conduct ablation studies on edge construction, know-aware attention, the number of neighbor nodes, convergence, and efficiency. Finally, we perform additional experiments on the in-house test set to show the generalization of the trained models.
 
\subsection{Datasets}
\begin{table}[htbp]
\centering
\resizebox{1\columnwidth}{!}{
\renewcommand{\arraystretch}{1.4}{
\begin{tabular}{c|c|cccc}
\hline
\multirow{2}{*}{\textbf{Datasets}} &
  \multirow{2}{*}{\textbf{Type}} &
  \multicolumn{4}{c}{\textbf{Number of WSIs}} \\ \cline{3-6} 
 &
   &
  \textbf{Stage \uppercase\expandafter{\romannumeral1}} &
  \textbf{Stage \uppercase\expandafter{\romannumeral2}} &
  \textbf{Stage \uppercase\expandafter{\romannumeral3}} &
  \textbf{Stage \uppercase\expandafter{\romannumeral4}} \\ \hline
\multirow{2}{*}{\textbf{TCGA-ESCA}}   & \textbf{1} & 30    & 56   & 75  & 19 \\
                                      & \textbf{2} & 13    & 115  & 55  & 12 \\ \hline
\multirow{3}{*}{\textbf{TCGA-KIDNEY}} & \textbf{1} & 129   & 109  & 62  & 26 \\
                                      & \textbf{2} & 270   & 58   & 124 & 83 \\
                                      & \textbf{3} & 231   & 44   & 70  & 27 \\ \hline
\multirow{2}{*}{\textbf{TCGA-LUNG}}   & \textbf{1} & 644   & 378  & 232 & 20 \\
                                      & \textbf{2} & 464   & 200  & 141 & 42 \\ \hline
\multirow{2}{*}{\textbf{FROZEN-LUNG}} & \textbf{1} & \multicolumn{4}{c}{65}  \\ \cline{3-6} 
                                      & \textbf{2} & \multicolumn{4}{c}{105} \\ \hline
\end{tabular}}}
\caption{Statistics of three public datasets and in-house datasets.}
\label{datasets}
\end{table}

We evaluate our WiKG and other methods on three public clinical WSI benchmark datasets from The Cancer Genome Atlas (TCGA) project \cite{weinstein2013cancer}, including ESCA, KIDNEY, and LUNG. We performed two-group classification experiments on tumor type and stage for each dataset. TCGA-ESCA comprises an esophageal cancer cohort of 375 cases, including adenocarcinomas (1) and squamous cell carcinomas (2). TCGA-KIDNEY comprises a kidney cancer cohort of 1,233 cases, including chromophobe renal cell carcinoma (1), renal clear cell carcinoma (2), and renal papillary cell carcinoma (3). TCGA-LUNG comprises a lung cancer cohort of 2121 cases, including squamous cell carcinomas (1) and adenocarcinomas (2). For staging tasks, all cases are divided into "\uppercase\expandafter{\romannumeral1}", "\uppercase\expandafter{\romannumeral2}", "\uppercase\expandafter{\romannumeral3}" and "\uppercase\expandafter{\romannumeral4}" categories according to TNM standards \cite{goldstraw2007iaslc}. 

For preprocessing, each WSI is divided into 256 $\times$ 256 non-overlapping patches at $20 \times$ magnifications. The Area Under the Curve (AUC) is reported for all experiments due to its comprehensiveness and insensitivity to class imbalance. Additionally, Accuracy and weighted F1-score are also reported. These metrics are determined by a threshold of 0.5. All results on TCGA datasets are obtained through 4-fold cross-validation, with each metric recorded as a percentage value accompanied by the standard error. We visualize the constructed graph structure in the Camelyon16 dataset \cite{bejnordi2017diagnostic}, chosen for its pixel-level annotations of tumor mask. To further evaluate the generalization performance of our proposed method, we also conduct in-house testing with 170 real cases of frozen section WSIs collected from the First Affiliated Hospital of Sun Yat-sen University, including squamous cell carcinomas (1) and adenocarcinomas (2), documented under the FROZEN-LUNG. Table \ref{datasets} shows more details of the dataset.

\begin{table*}[tbp]
\centering
\resizebox{2\columnwidth}{!}{
\renewcommand{\arraystretch}{1.5}{
\begin{tabular}{lccccccccc}
\hline
\multirow{2}{*}{\textbf{Method}} &
  \multicolumn{3}{c}{\textbf{TCGA-ESCA}} &
  \multicolumn{3}{c}{\textbf{TCGA-KIDNEY}} &
  \multicolumn{3}{c}{\textbf{TCGA-LUNG}} \\ \cline{2-10} 
 &
  \textbf{Accuracy} &
  \textbf{AUC} &
  \textbf{F1-score} &
  \textbf{Accuracy} &
  \textbf{AUC} &
  \textbf{F1-score} &
  \textbf{Accuracy} &
  \textbf{AUC} &
  \textbf{F1-score} \\ \hline
ABMIL \cite{ilse2018attention} &
  $51.21_{4.10}$ &
  $64.28_{4.00}$ &
  \multicolumn{1}{c|}{$48.51_{2.75}$} &
  $51.99_{2.08}$ &
  $63.42_{2.24}$ &
  \multicolumn{1}{c|}{$47.10_{2.45}$} &
  $52.57_{0.72}$ &
  $57.33_{1.42}$ &
  $45.39_{0.98}$ \\
CLAM-SB \cite{lu2021data} &
  $51.47_{3.24}$ &
  $65.08_{1.46}$ &
  \multicolumn{1}{c|}{$48.70_{3.84}$} &
  $51.99_{2.62}$ &
  $66.07_{2.97}$ &
  \multicolumn{1}{c|}{$47.83_{2.44}$} &
  $51.02_{2.19}$ &
  $57.22_{2.86}$ &
  $44.73_{2.95}$ \\
CLAM-MB \cite{lu2021data} &
  $54.15_{4.65}$ &
  $65.81_{2.30}$ &
  \multicolumn{1}{c|}{$51.56_{3.76}$} &
  $51.91_{3.27}$ &
  $67.67_{2.50}$ &
  \multicolumn{1}{c|}{$48.63_{1.85}$} &
  $50.21_{1.66}$ &
  $57.32_{2.01}$ &
  $43.99_{1.52}$ \\
DSMIL \cite{li2021dual} &
  $51.75_{4.58}$ &
  $66.03_{6.05}$ &
  \multicolumn{1}{c|}{$49.07_{5.85}$} &
  $52.32_{3.02}$ &
  $66.85_{2.80}$ &
  \multicolumn{1}{c|}{$48.16_{2.29}$} &
   $52.33_{1.21}$&		
   $57.93_{2.74}$&
   $44.96_{2.07}$\\
TransMIL \cite{shao2021transmil} &
  $53.61_{3.44}$ &
  $65.43_{4.07}$ &
  \multicolumn{1}{c|}{$49.86_{5.96}$} &
  $50.37_{1.43}$ &
  $61.96_{2.64}$ &
  \multicolumn{1}{c|}{$43.73_{2.60}$} &
   $52.15_{1.06}$&		
   $55.09_{0.94}$&
   $42.51_{1.49}$\\
DTFD-MIL \cite{zhang2022dtfd} &
  $50.68_{5.29}$ &
  $65.60_{5.18}$ &
  \multicolumn{1}{c|}{$48.21_{5.26}$} &
  $51.75_{2.00}$ &
  $64.03_{4.26}$ &
  \multicolumn{1}{c|}{$45.63_{3.38}$} &
  $52.52_{0.78}$ &		 
  $57.56_{1.71}$ &
  $44.88_{1.76}$\\
HIPT \cite{chen2022scaling} &
  $48.82_{6.16}$ &
  $63.46_{3.89}$ &
  \multicolumn{1}{c|}{$35.55_{12.13}$} &
  $51.81_{2.96}$ &
  $61.83_{1.40}$ &
  \multicolumn{1}{c|}{$37.40_{1.53}$} &
  $52.24_{4.18}$ & 
  $55.26_{0.46}$ &
  $35.92_{4.74}$ \\
GTP \cite{zheng2022graph} &
  $46.38_{4.03}$ &
  $63.55_{3.88}$ &
  \multicolumn{1}{c|}{$45.90_{3.96}$} &
  $46.05_{2.14}$ &
  $57.01_{1.00}$ &
  \multicolumn{1}{c|}{$39.73_{2.14}$} &
  $50.85_{3.77}$ & 
  $55.22_{4.10}$ &
  $43.34_{2.12}$ \\
Patch-GCN \cite{chen2021whole} &
  $52.28_{3.98}$ &
  $68.03_{3.72}$ &
  \multicolumn{1}{c|}{$50.47_{5.18}$} &
  $54.26_{3.26}$ &
  $68.96_{2.22}$ &
  \multicolumn{1}{c|}{$49.89_{2.26}$} &
  $50.63_{0.58}$ &          		
  $53.82_{1.43}$ &
  $42.62_{2.64}$ \\ \hline
\textbf{WiKG (ours)} &
  $\mathbf{57.50_{3.99}}$ &
  $\mathbf{69.96_{4.28}}$ &
  \multicolumn{1}{c|}{$\mathbf{55.80_{4.39}}$} &
  $\mathbf{55.49_{2.12}}$ &
  $\mathbf{69.71_{1.40}}$ &
  \multicolumn{1}{c|}{$\mathbf{51.23_{1.43}}$} &
   $\mathbf{52.85_{0.74}}$ &      		
   $\mathbf{60.34_{1.37}}$ &
   $\mathbf{47.52_{1.52}}$\\ \hline
\end{tabular}}}
\caption{Cancer staging results [\%] of various methods on TCGA-ESCA, TCGA-KIDNEY, and TCGA-LUNG datasets.}
\label{staging}
\end{table*}

\subsection{Implementation Details}
The proposed methodology uses the Pytorch library in Python on an Nvidia RTX 3090 GPU. To adopt a single GPU and facilitate future work, we utilize a ViT Small model \cite{dosovitskiy2020image} pretrained on ImageNet \cite{deng2009imagenet} as the encoder. For comparative experiments utilizing domain-specific encoders trained on pathology images, please refer to the Supplementary material. We extract a 384-dimensional feature vector for each patch and then expand it to 512 dimensions via a fully connected layer, forming the final feature representation of the patch. The batch size is set to 1, indicating that only one slide is processed per iteration. The $k$ in Top-k is chosen to be 6, which means that each patch will have 6 patches connected to it. We set the number of epochs to 100 and employ the Adam \cite{kingma2014adam} optimizer with a learning rate of $10^{-4}$ and a weight decay of $10^{-5}$ for model weight updates. A dropout ratio of 0.3 is applied before obtaining graph-level embeddings. All other baseline methods are experimented with the same parameters and settings.

\subsection{Comparison with Other Existing Works}

We present our tumor typing and staging results on three datasets in Table \ref{typing} and Table \ref{staging}, respectively, and compare them with the following nine state-of-the-art methodologies: (1) ABMIL \cite{ilse2018attention}, a MIL framework that aggregates instance information via an attention mechanism to obtain bag-level embeddings. (2) CLAM-SB \cite{lu2021data}, a single-gated attention-based MIL framework optimized with clustering constraint loss. (3) CLAM-MB \cite{lu2021data}, an enhanced version of CLAM-SB employing multi-gated attention. (4) DSMIL \cite{li2021dual}, a dual-stream MIL method aggregating patch embeddings through non-local attention pooling and max pooling. (5) TransMIL \cite{shao2021transmil}, a MIL-like paradigm using transformers and multiscale position encoding modules. (6) DTFD-MIL \cite{zhang2022dtfd}, a double-tier MIL framework developed by introducing pseudo-bags and feature distillation. (7) HIPT \cite{chen2022scaling}, a hierarchical ViT architecture for leveraging Pyramid WSI structure. (8) GTP \cite{zheng2022graph}, a graph representation integrating graph and transformer. (9) Patch-GCN \cite{chen2021whole}, a hierarchical graph-based method for WSI survival prediction, with global attention pooling introduced as a graph representation model. Our method demonstrates superior performance compared to MIL and graph representation methods, suggesting that our WiKG might better capture the interactions between patches, thus obtaining more comprehensive global information. Notably, in staging tasks, the improvement in F1-score is more significant than in Accuracy, indicating the advantage of WiKG in delineating the malignancy degree of subdivided tumors. It is necessary to acknowledge that the staging of WSIs is generally influenced by factors including the type of tumor, its dimension, the extent of invasion into adjacent organs, the number of cancerous cells in lymph nodes, and the presence of distant metastases. Most current algorithms employed in this domain are adaptations from those developed for typing tasks. In our future work, we will explore deep learning methods for tumor staging.

\subsection{Effectiveness of Proposed Edge Construction}
\begin{table}[htbp]
\centering
\resizebox{1\columnwidth}{!}{
\renewcommand{\arraystretch}{1.3}{
\begin{tabular}{llccc}
\hline
\multirow{2}{*}{\textbf{Datasets}} & \multirow{2}{*}{\textbf{Task}} & \multicolumn{3}{c}{\textbf{Edge Construction}} \\ \cline{3-5} 
            &  & \textbf{k-NN (cos)}    & \textbf{k-NN (dist)}  & \textbf{WiKG (ours)} \\ \hline
ESCA   & Type  & $94.98_{2.52}$ & $92.36_{1.11}$ & $\mathbf{95.23_{2.90}}$       \\
       & Stage & $64.36_{4.38}$ & $66.27_{4.69}$ & $\mathbf{69.96_{4.28}}$       \\
KIDNEY & Type  & $99.44_{0.28}$ & $99.47_{0.24}$ & $\mathbf{99.65_{0.12}}$       \\
       & Stage & $68.62_{1.97}$ & $68.01_{1.77}$ & $\mathbf{69.71_{1.40}}$       \\
LUNG   & Type  & $90.62_{1.40}$ & $88.31_{2.22}$ & $\mathbf{90.78_{1.24}}$       \\
       & Stage & $58.72_{1.58}$ & $57.92_{1.99}$ & $\mathbf{60.34_{1.37}}$       \\ \hline
\end{tabular}}} 
\caption{Cancer typing and staging AUC results [\%] of our proposed dynamic graph construction compared to graph construction based on the k-nearest neighbor algorithm.}
\label{knn}
\end{table}

Our strategy to construct directed edges based on the heads and tails relation demonstrates efficiency in establishing topological structures. In Table \ref{knn}, we report a comparative analysis of the AUC performance between our method and two conventional dynamic edge construction algorithms based on the k-nearest neighbor (k-NN). Specifically, k-NN (dist) and k-NN (cos) identify the top k indexes with the highest similarity by calculating the Euclidean distance and cosine similarity between feature embeddings respectively, and keep other structures unchanged. Notably, our approach outperforms the other two k-NN-based methods by nearly one to three percentage points across two WSI analysis tasks in all three datasets, suggesting that our edge-building strategy may be more consistent with representations of the organizational microenvironment.

%Notably, our approach enhances nearly one to three percentage points across two WSI analysis tasks in all three datasets, suggesting that our edge-building strategy may be more consistent with representations of the organizational microenvironment. %Specifically, the k-NN edge construction strategy calculates the Euclidean distance between feature embeddings to identify the top-k most similar indices as neighbor nodes and keep other structures unchanged.

\subsection{Effectiveness of Knowledge-aware Attention}

\begin{table}[htbp]
\centering
\resizebox{1\columnwidth}{!}{
\renewcommand{\arraystretch}{1.3}{
\begin{tabular}{lccc}
\hline
\textbf{GNN} & \textbf{ESCA}      & \textbf{KIDNEY}    & \textbf{LUNG}      \\ \hline
GCN \cite{kipf2016semi}  & $94.24_{2.84}$ & $99.44_{0.14}$ & $90.33_{1.20}$ \\
GIN \cite{xu2018powerful}  & $93.58_{2.37}$ & $99.56_{0.18}$ & $90.37_{1.32}$ \\
SAGE \cite{hamilton2017inductive} & $93.52_{2.60}$ & $99.51_{0.23}$ & $90.40_{1.14}$ \\
GAT \cite{velivckovic2018graph}  & $93.77_{2.68}$ & $99.54_{0.21}$ & $90.41_{1.11}$ \\ \hline
\textbf{WiKG(ours)}       & $\mathbf{95.23_{2.90}}$ & $\mathbf{99.65_{0.12}}$ & $\mathbf{90.78_{1.24}}$ \\ \hline
\end{tabular}}}
\caption{Cancer typing AUC results [\%] of our proposed WiKG compared to other GNN architecture.}
\label{gnn}
\end{table}

We conduct a comparative performance analysis with other prevalent GNN architectures to further substantiate the performance of WiKG. We substituted our knowledge-aware attention mechanism update strategy with four different GNN architectures, which are (1) graph convolutional network (GCN) \cite{kipf2016semi}, graph isomorphism network (GIN) \cite{xu2018powerful}, graphSage (SAGE) \cite{hamilton2017inductive}, and graph attention network (GAT) \cite{velivckovic2018graph}. All other structures of the model remain unchanged. Table \ref{gnn} shows the performance comparison of our proposed WiKG with different GNN architectures, demonstrating the superiority of our proposed method in terms of AUC across three WSI typing tasks. Especially in the cancer analysis of esophageal, the proposed method improves up to nearly 2\%. It is worth noting that GAT is also a GNN framework that aggregates node features through attention scores. Still, its performance is not as good as WiKG, which shows that WiKG may have more advantages in capturing effective information of neighbors. %For discussion of cancer staging results, please refer to the Supplementary material.

\subsection{Effect of Numbers of Neighbor Nodes}
\begin{figure}[t]
\centerline{\includegraphics[width=1\columnwidth]{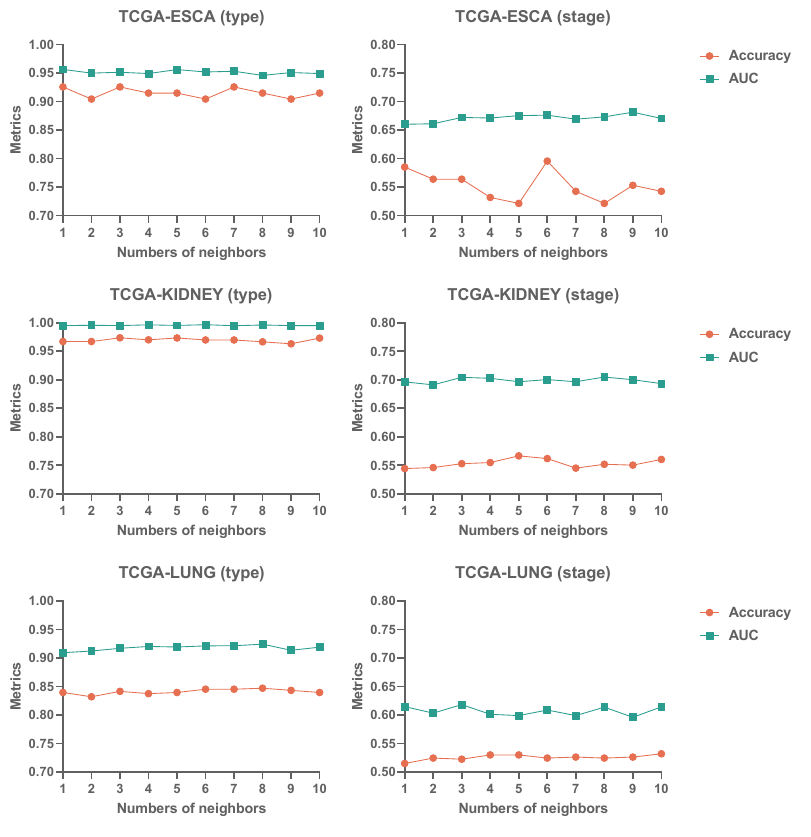}}
\caption{Typing and staging results of AUC and Accuracy scores with different numbers of neighbor nodes on three TCGA datasets.}
\label{ablation}
\end{figure}

We also conduct analysis experiments on the different numbers of neighbor nodes. Figure \ref{ablation} illustrates the Accuracy and AUC scores of WiKG in the typing and staging experiments across three TCGA datasets with varying numbers of neighbor nodes. Overall, the number of neighbor nodes has little impact on the performance of WiKG. Although the Accuracy metric fluctuates in the staging experiment of TCGA-ESCA, this fluctuation is reasonable considering the uneven number of each staging. At the same time, this may also be related to the inherent morphological diversity and complexity of esophageal cancer.

\subsection{Convergence and Time Efficiency}
\begin{figure}[t]
\centerline{\includegraphics[width=1\columnwidth]{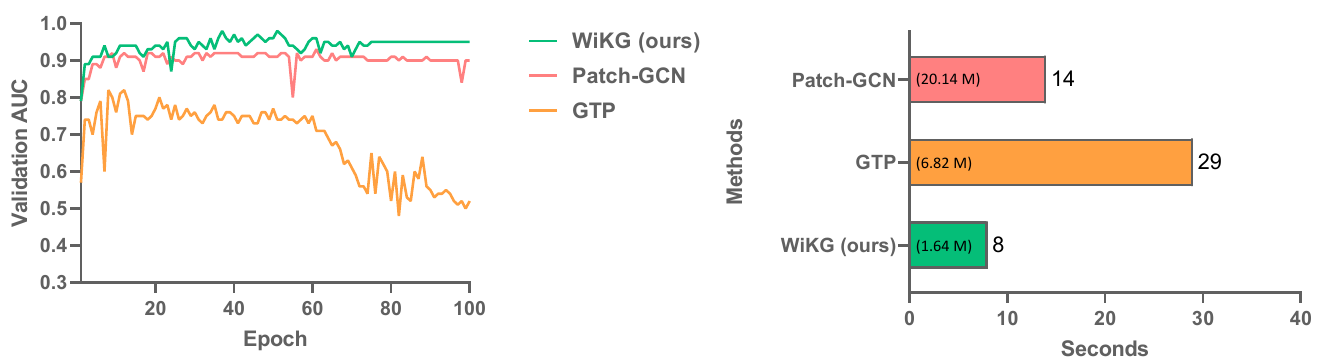}}
\caption{Convergence curves of validation AUC and training time of each epoch in TCGA-ESCA.}
\label{times}
\end{figure}

The over-parameterization problem can lead to unstable convergence of GNNs and increased training time. We compare WiKG with Patch-GCN and GTP in the staging task of TCGA-ESCA. Figure \ref{times} compares our convergence curve and the training time of each epoch. WiKG converges quickly during training and shows no over-fitting phenomenon compared with Patch-GCN and GTP. At the same time, in terms of training speed in a single epoch, WiKG is also the fastest and has the smallest number of parameters.

\subsection{Visualization and Further Experiment}
\begin{figure}[tbp]
\centerline{\includegraphics[width=0.9\columnwidth]{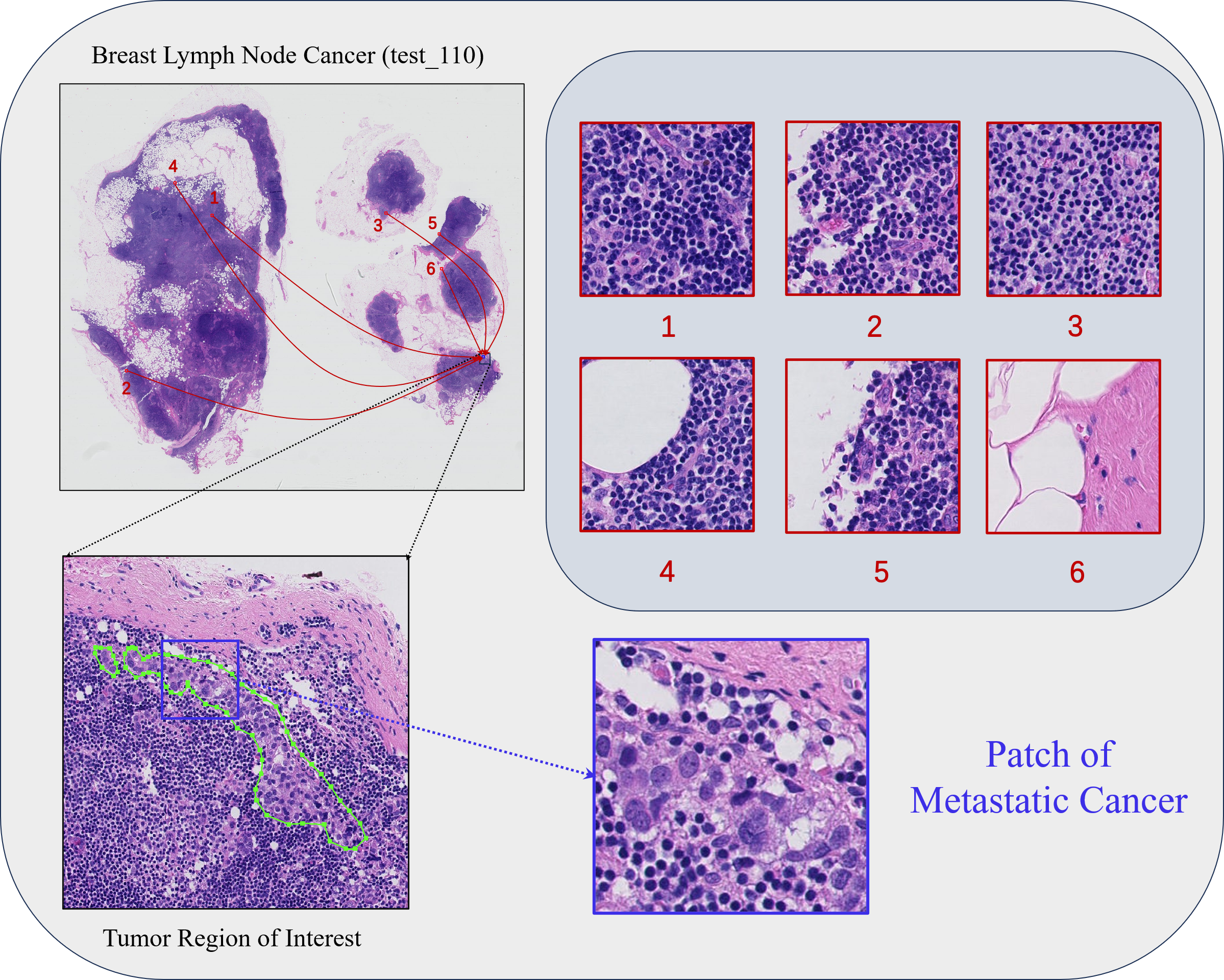}}
\caption{Visualization of the constructed graph structure.}
\label{visual}
\end{figure}

Through the visualization of the directed graph structure constructed by WiKG, we gain an intuitive understanding of the potential relationships between patches, further helping to build relationship networks of cancer cells within tissues and enriching the representation of the tumor microenvironment. Figure \ref{visual} displays our visualization results on the Camelyon16 dataset. We observe that the patch with metastatic cancer is more likely to select patches containing a large number of lymphocytes as neighbors and some adipose and fibrous tissue as supplements. Notably, the patch with the highest knowledge-aware attention score (number $5$) contains a greater concentration of lymphomas, indicating that the patch containing metastatic cancer pays more attention to patches containing more lesion features. 

\begin{figure}[tbp]
\centerline{\includegraphics[width=1\columnwidth]{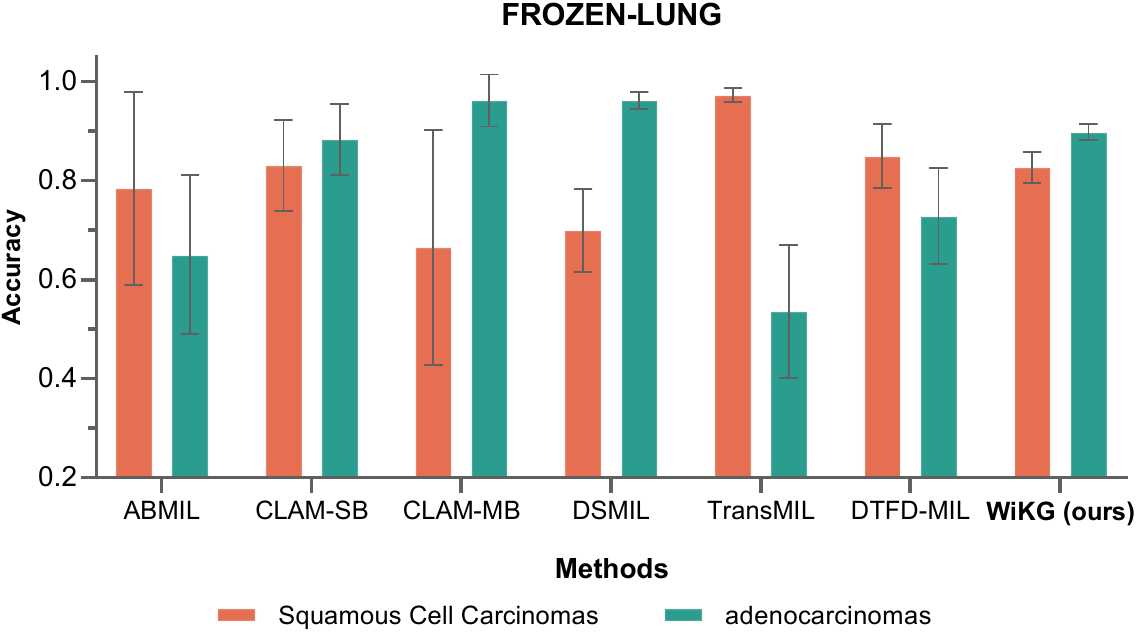}}
\caption{Mean and standard deviation of per-class Accuracy with various methods on FROZEN-LUNG datasets.}
\label{luadvslusc}
\end{figure}

We also conduct further testing experiments on the FROZEN-LUNG dataset with models previously trained well on the TCGA-LUNG dataset. The purpose of these experiments is to evaluate the generalization performance and application prospects of various methods. Table \ref{frozen-lung} provides a comparative analysis with other methods, demonstrating the superior overall performance of WiKG. At the same time, we also show the results of the Accuracy metric for each category, which is demonstrated in Figure \ref{luadvslusc}. Specifically, WiKG has excellent recognition capabilities for both lung subtypes, and the standard deviation of each category is also much smaller than other methods. However, the generalization capabilities of other MIL methods are not comparable to WiKG.

\begin{table}[htbp]
\centering
\resizebox{1\columnwidth}{!}{
\renewcommand{\arraystretch}{1.2}{
\begin{tabular}{lccc}
\hline
\multirow{2}{*}{\textbf{Method}} & \multicolumn{3}{c}{\textbf{FROZEN-LUNG}}                                    \\ \cline{2-4} 
           & \textbf{Accuracy} & \textbf{AUC}   & \textbf{F1-score} \\ \hline
ABMIL \cite{ilse2018attention}    & $70.15_{7.15}$    & $80.12_{10.73}$ & $70.73_{8.41}$    \\ 
CLAM-SB \cite{lu2021data}  & $84.71_{2.40}$    & $92.09_{3.11}$ & $84.41_{2.43}$    \\ 
CLAM-MB \cite{lu2021data}  & $83.82_{4.91}$    & $93.71_{0.98}$ & $83.64_{4.91}$    \\ 
DSMIL \cite{li2021dual}    & $86.18_{2.61}$    & $93.91_{0.29}$ & $85.66_{2.90}$    \\ 
TransMIL \cite{shao2021transmil} & $70.29_{8.07}$    & $\mathbf{94.10_{1.66}}$ & $69.64_{8.91}$    \\ 
DTFD-MIL \cite{zhang2022dtfd} & $77.50_{4.25}$    & $87.28_{3.46}$ & $77.70_{4.21}$    \\ \hline 
\textbf{WiKG (ours)}             & $\mathbf{87.06_{1.66}}$ & $92.31_{0.93}$ & $\mathbf{87.05_{1.68}}$ \\ \hline
\end{tabular}}}
\caption{Results [\%] of various methods on FROZEN-LUNG datasets. All methods are trained well on TCGA-LUNG datasets.}
\label{frozen-lung}
\end{table}
Some interesting results are observed in this experiment. For most methods, there is a significant difference between AUC and accuracy/F1-score, especially for TransMIL. Although its Accuracy and F1-score are relatively low, its AUC is unusually high, specifically for squamous cell carcinomas, which is nearly 44 percent Accuracy higher than that of adenocarcinomas. This may be due to the imbalance in the size of the training data, which is that the greater the number of a specific category in the training set, the higher the Accuracy of its test. However, other methods show the opposite results except TransMIL, ABMIL, and DTFD-MIL. We also observe that other compared methods typically encountered two issues: a substantial discrepancy in Accuracy across different categories and a high standard deviation in Accuracy of specific categories. This suggests that some existing works might suffer from overfitting issues and exhibit increased sensitivity to particular categories. Our proposed WiKG, in contrast, significantly mitigated both these issues.

\section{Conclusion and Future Work}
In this paper, we introduce a novel dynamic graph representation approach called WiKG to analyze WSIs. By modeling interactions between head and tail embeddings and constructing a knowledge-aware attention mechanism that aggregates neighbor information, WiKG liberates the ability of patches to explore mutual relationships in their topological structures and also uses the directional contributions between entities to better interact between patches, thereby achieving better WSI analysis performance. Through extensive experiments and ablation studies of three public benchmark datasets and in-house test sets, we demonstrate the effectiveness and better generalization performance of WiKG. 

In future works, we will focus on the interpretability of WiKG in WSIs and explore the impact of graph pooling on WSIs. We will also conduct in-depth research on the generalization of WSI analysis tasks, because existing research methods may not have good generalization performance.
% In our future work, we will focus on the interpretability of dynamic graphs in WSIs and explore the impact of graph pooling on WSIs. We will also conduct in-depth research on the generalization of WSI analysis tasks, because existing research methods may not have good generalization performance.We will also conduct in-depth research on the generalization of WSI analysis tasks because existing research methods may not have good generalization performance, which will have a significant impact on the actual deployment and application of the algorithm.

% \noindent\textbf{Acknowledgement:} This work was supported by the Component Project of Shenzhen Pathology Medical Imaging Intelligent Diagnosis Engineering Research Center (XMHT20230115004), Science and Technology Foundation of Shenzhen City (KCXFZ202012211173207022).

% The work was supported in part by XMHT20230115004 and KCXFZ20201221173207022, the Development and Reform Commission of Shenzhen Municipality. The authors declare that they have no known competing financial interests or personal relationships that could have appeared to influence the work reported in this paper.

{
    \small
    \bibliographystyle{ieeenat_fullname}
    \bibliography{main}
}


\end{document}